\documentclass[first]{iscram}

\usepackage[utf8]{inputenc}
\usepackage[style=apa]{biblatex} 
\usepackage{todonotes}
\usepackage{subfig}
\usepackage{booktabs}
\usepackage{hyperref}
\usepackage{tikz}
\usepackage{fancyvrb}
\usepackage{listings}
\usepackage{enumitem}
\usepackage{tcolorbox}
\usepackage{multicol}
\usepackage{siunitx}
\usepackage{xpatch}
\usepackage{cleveref}
\lstdefinestyle{common}{
  xleftmargin=.5em,
  xrightmargin=.5em,
  frame=single,framesep=.5em,framerule=0pt,
  fancyvrb=true,
  basicstyle=\ttfamily,
  keywordstyle=\color{cyan!50!blue!75!black}\bfseries,
  commentstyle=\color{red!50!black}\itshape,
  stringstyle=\ttfamily\color{green!50!black},
  numbers=none,
  showspaces=false,
  showstringspaces=false,
  fontadjust=true,
  keepspaces=true,
  flexiblecolumns=true,
  emphstyle=\color{red},
}
\lstdefinestyle{TeX}{
  style=common,
  backgroundcolor=\color{blue!5},
  aboveskip=5pt,
  belowskip=5pt,
  language=[LaTeX]TeX,
  moretexcs={
    abstract, addbibresource, iscramset, keywords, mainmatter,
    maketitle, printbibliography, subsection, subsubsection, url,
    urldef, href, includegraphics, ldots, parencite, citeauthor,
    citeyear, citetitle, midrule, toprule, bottomrule
  },
  fancyvrb=true,
}
\lstdefinestyle{console}{
  style=common,
  backgroundcolor=\color{gray!10},
  aboveskip=5pt,
  belowskip=5pt,
}

\newlist{options}{description}{1}
\setlist[options]{%
  beginpenalty=10000,%
  itemsep=.5\parskip plus .3\parskip minus .2\parskip,
  parsep=.5\parskip plus .3\parskip minus .2\parskip,
  topsep=.5\parskip plus .3\parskip minus .2\parskip,
  partopsep=.5\parskip plus .3\parskip minus .2\parskip,
  style=nextline,labelindent=1em,%
  font=\normalfont\ttfamily}

\colorlet{macro color}{cyan!50!blue!75!black}
\colorlet{option color}{red!50!black}
\colorlet{generic color}{green!40!black}

\newtcolorbox{pseudoTeX}{colback=blue!5,colframe=blue!5,before=\nobreak}
\let\LaTeXorig\LaTeX
\renewcommand\LaTeX{\bgroup\fontfamily{lmr}\selectfont\upshape\LaTeXorig\egroup}

\urldef{\sitewebgind}\url{http://gind.mines-albi.fr}
\urldef{\sitewebminesalbi}\url{http://www.mines-albi.fr}
\urldef{\jdmail}\url{...@example.com}
\urldef{\sitex}\url{www.example.com}

\iscramset{
  CoRe Paper 2024={Geographic Information Science (GIS) for Crisis Management},
  title={
    QuakeSet: A Dataset and Low-Resource Models to Monitor Earthquakes through Sentinel-1
  },
  short title={QuakeSet},
  author={
    short name={Daniele Rege Cambrin},
    full name=Daniele Rege Cambrin\thanks{corresponding author},
    affiliation={
      Politecnico di Torino\\
       \href{mailto:daniele.regecambrin@polito.it}{\url{daniele.regecambrin@polito.it}}
    },
  },
  author={
    full name= Paolo Garza,
    affiliation={
      Politecnico di Torino\\
       \href{mailto:paolo.garza@polito.it}{\url{paolo.garza@polito.it}}
    },
  },
}

\usepackage{tikzpagenodes}
\usetikzlibrary{fit}

\begin{document}

\maketitle

\abstract{Earthquake monitoring is necessary to promptly identify the affected areas, the severity of the events, and, finally, to estimate damages and plan the actions needed for the restoration process. The use of seismic stations to monitor the strength and origin of earthquakes is limited when dealing with remote areas (we cannot have global capillary coverage). Identification and analysis of all affected areas is mandatory to support areas not monitored by traditional stations. Using social media images in crisis management has proven effective in various situations. However, they are still limited by the possibility of using communication infrastructures in case of an earthquake and by the presence of people in the area. Moreover, social media images and messages cannot be used to estimate the actual severity of earthquakes and their characteristics effectively. The employment of satellites to monitor changes around the globe grants the possibility of exploiting instrumentation that is not limited by the visible spectrum, the presence of land infrastructures, and people in the affected areas. In this work, we propose a new dataset composed of 
images taken from Sentinel-1 and a new series of tasks to help monitor earthquakes from a new detailed view. Coupled with the data, we provide a series of traditional machine learning and deep learning models as baselines to assess the effectiveness of ML-based models in earthquake analysis.}


\section{Introduction}
The application of modern machine learning and deep learning solutions to remote sensing is strictly linked to data availability. Thanks to the launch of Landsat-8 \parencite{landsat}, MODIS \parencite{modis}, and Sentinel missions \parencite{sentinel1,sentinel2}, general-purpose datasets were released in recent years \parencite{ssl4eo,eurosat,bigearthnet}. Emergency management can benefit from these solutions. 
However, only a few datasets of satellite imagery were publicly released for a limited number of emergency management tasks (e.g., flood detection \parencite{sen1floods} and burned areas delineation \parencite{cabuar}). Ad-hoc image datasets are unavailable for many emergency management tasks.
Among the others, public datasets of satellite imagery tailored to training machine learning models for earthquake analysis have not been released to our knowledge.

Earthquake monitoring is mainly based on seismometers that actively record seismic waves to detect, understand the strength, and identify the origin of these events. However, we cannot have global capillary coverage using this instrumentation.
Conversely, satellite instruments have a broader view over the globe. Hence, they can help reach remote places worldwide and understand which areas were affected by the earthquakes without sending people to assess the damages and without using seismometers.
Tasks such as identifying affected areas and estimating magnitude can be solved offline by collecting satellite images and then applying machine learning models in a data center. Alternatively, the inference can be made directly onboard the satellites using low-resource models, avoiding storing large amounts of data and enabling the possibility of transmitting only a few significant information (e.g., coordinates of the damaged area, the estimated magnitude, etc).
This raises another crucial challenge linked to the resource availability of the used systems: employing a large model requires computational resources unavailable on board. Creating an accurate model is not the only requirement. Low resource consumption is mandatory, and memory, CPU, and energy must be used effectively.

In this context, we propose a new dataset of more than a hundred earthquakes around the Earth, containing thousands of tri-temporal samples of the same event using Sentinel-1 data and ISC annotations. Moreover, we assess the effectiveness of shallow and deep learning models to solve tasks related to earthquake monitoring.

Our contributions can be summarized as follows:
\begin{itemize}
    \item We propose a series of tasks to analyze earthquake-affected areas that can be solved by applying machine learning solutions to Sentinel-1 data.
    \item We evaluate a series of deep learning and shallow machine learning solutions in a subset of the proposed tasks.
    \item We pose the tasks as a balance between performance and resource consumption to democratize the proposed solutions and make feasible the application to low-resource devices.
    \item We publicly release a collection composed of tri-temporal time-series images, allowing the possibility to analyze time-related changes.
\end{itemize}

For reproducibility, we released the code for the experiments at \url{https://github.com/DarthReca/quakeset/}. The dataset is available at \url{https://huggingface.co/datasets/DarthReca/quakeset} to allow the community to experiment with it.

\section{Related Works}

\subsection{Image processing through deep learning}
The rise of convolutional neural networks and, more recently, vision transformers significantly improved the capability of processing images. The great data availability and the use of pre-trained neural networks posed deep learning models as state-of-the-art solutions in computer vision over classical machine learning models in benchmark datasets like ImageNet \parencite{imagenet}. The increasing performance of these models is often linked to the increased computational costs at training time and in inference \parencite{segformer}. Dealing with real-time constraints or low-resource devices poses the challenge of balancing accuracy with resource consumption. Developing specific deep learning models like MobileNets \parencite{mobilenetv1,mobilenetv2} or MobileViTs \cite{mobilevit,mobilevitv2} were strictly linked to these constraints. Other solutions were designed in different fashions, including the possibility of employing less accurate real-time versions or the heavier and best-performing ones \parencite{segformer,convnext,convnetxv1,pyramidvit}.
Finally, the effectiveness of the deep learning models in computer vision is highly impacted by data availability. Without ad hoc collections of images, the results decrease rapidly when considering domains for which tailored datasets are unavailable or limited.

\subsection{Image processing and deep learning in crisis management}
Using textual information to analyze the evolution of ongoing emergencies has proven effective with the solutions proposed in the Incident Stream \parencite{istrack} and Crisis Facts \parencite{crisisfacts} tracks of TREC. Although, also the analysis of images taken from social media has proven to be effective in the identification of areas affected by fires \parencite{fire_social}, earthquakes, hurricanes, and typhoons \parencite{damage_social, flood_social}. 
The training of deep convolutional neural networks grants improved results over traditional techniques like Bag-Of-Visual-Words \parencite{damage_social}.
The release of Incidents1M \parencite{incidents1m} with around a million images of 43 incidents granted the possibility to pre-train vision transformers and finetune existing pre-trained ones \parencite{crisisvit}. 
The significant drawback of all existing solutions is that they rely on social media and RGB images. 
RGB images can limit the comprehension of certain aspects of the events due to the limitations of the visible spectrum. For instance, using the infrared spectrum in Sentinel-2 helps monitor green areas \parencite{sentinel2_overview}, being vegetation more sensitive to these frequencies. Moreover, using social media data also excludes the possibility of reaching sites with limited human presence or low-quality or damaged communications since they rely on information people send over social media.
For these reasons, we propose to use satellite data, specifically Sentinel-1 images. 

\subsection{Machine Learning in Remote Sensing}
The advancement of neural networks greatly benefits applications in the remote sensing field \parencite{remote_sensing_review,remote_sensing_review2}. The networks designed for RGB images have proven adaptable to other spectral bands like infrared and ultra blue. An extensive series of classical computer vision tasks, such as image classification and segmentation, were adapted to the remote sensing domain. The BigEarthNet \parencite{bigearthnet} and EuroSAT \parencite{eurosat} datasets pose the problem of landcover detection as an image classification one. The Seaships dataset \parencite{ship_detection} analyzes an object detection problem, while CaBuAr \parencite{cabuar} analyzes a semantic segmentation problem in the burned area identification context. 
In addition to the extra spectral bands per image, in many cases, these datasets contain spatiotemporal information granting the possibility of solving tasks like change detection, which aims to find in supervised or unsupervised way the difference between two samples of the same spatial area taken at two different timestamps \parencite{change_detection_survey}. The WHU dataset \parencite{urban_change_detection} deals with this challenge for urban planning.

\subsection{Machine Learning in Seismology}
There is an increasing interest in applying machine learning and deep learning in the seismology field \parencite{magnitude_estimation_review}. The current machine-learning solutions were mainly applied to earthquake seismic waves (P-waves and S-waves) to detect an earthquake in real-time. Machine learning was also applied to distinguish between earthquakes and microtremors \parencite{earthquake_microtremors} and to phase picking \parencite{earthquake_transformers,deep_learning_pwaves}. Current solutions involve convolution neural networks \parencite{p_detector,deep_learning_pwaves}, generative adversarial networks \parencite{earthquake_microtremors}, and transformers \parencite{earthquake_transformers} applied to seismic waves. Sentinel-1 imagery has already been tested in earthquake analysis without involving automatic systems \parencite{sentinel1_earthquakes}. Specifically, the reported manual analysis highlights the feasibility of earthquake detection using Sentinel-1 data. The authors of \cite{sentinel1_earthquakes} managed to identify approximately $70\%$ of considered earthquakes. Compared to the manual analysis, we do not need to apply preprocessing steps for water bodies or atmospheric correction, and, more importantly, we propose completely automated ML-based approaches. Nevertheless, manual analysis in \cite{sentinel1_earthquakes} involves correlation, which fails to understand non-linear relationships in data, while in our study, we employed many non-linear estimators. 
Moreover, we collected a time series of images to overcome the possible limitation of single image analysis, opening the possibility of leveraging changes instead of simply analyzing the content of single post-event images. Another difference is that we propose automatic solutions to analyze earthquakes from Sentinel-1 images instead of employing waves or manual solutions. Finally, we address several tasks, including detection and magnitude estimation.

\section{Sources Description}
In this section, we present the sources of our dataset, which includes Sentinel-1 GRD products and ground truth annotations from the International Seismological Centre (ISC). 

\subsection{Sentinel-1 mission}
Sentinel-1 \parencite{sentinel1}, with its Synthetic Aperture Radar (SAR), proved its effectiveness in a wide range of applications, and it is well suited for monitoring land changes due to its high revisit time and the capacity to see through clouds. It uses a single C-band operating at a center frequency of 5.405 GHz. 
SAR supports different acquisition modes in single and dual polarization: Stripmap (SM), Interferometric Wide swath (IW), Extra-Wide swath (EW), and Wave (WV). IW mode is mainly employed in land areas, WV in open oceans, EW in coastal areas, and SM for small islands. This work focused on land earthquakes, so we employed the IW mode, which provides VV and VH polarizations. Sentinel-1 products are available in four different formats: Level-0 Raw, Level-1 Ground Range Detected (GRD), Level-1 Single Look Complex (SLC), and Level-2 Ocean (OCN). We employed the Level-1 GRD products at two resolutions: $10 \times 10$ (high) and $40 \times 40$ (medium). To get all possible information, we kept only high-resolution samples.

\subsection{ISC Bulletin}
The International Seismological Centre (ISC) provides access to a database of all known and suspected earthquakes from more than 130 seismological agencies around the world \parencite{isc_data_collection}. The ISC Bulletin \parencite{isc_product} contains data from 1900 to the present (2023-11-19). The Reviewed ISC Bulletin, manually checked by ISC analysts and relocated (when there is sufficient data), is typically 24 months behind real-time and is currently up to 2021-11-01. The review procedure is applied to all earthquakes with unknown magnitude or greater than $3.5$. The procedure ensures the hypocenters are coherent with the origin region and that any regional and magnitude data are not missing. According to ISC, around $20\%$ of the entire database is reviewed.
The bulletin provides various information regarding each earthquake, such as hypocenter \parencite{isc_location}, the timestamp of the event, and magnitude \parencite{isc_magnitude}. Using this information, we can select an area of interest, know when to sample data related or not related to earthquakes, and try to understand different aspects of the events. 
The magnitude scale employed in this analysis and the ones directly provided by ISC is the mb scale, which can be used for medium entity earthquakes without encountering saturated values \parencite{magnitude_saturation}.
The following section describes how we use them to create tri-temporal time-series and ground truth annotations.

\section{Dataset}

\begin{table}[htb]
    \centering
    \caption{Dataset Summary}
    \label{tab:dataset_summary}
    \begin{tabular}{l|c}
         \hline
         Earthquakes & 155 \\\hline
         Temporal window & 2018-2021 \\\hline
         Image channels & 2 (VV and VH) \\\hline
         Area size & $20km \times 20km$ \\\hline
         Temporal difference & 1-13 days \\\hline
         Time series length & 3 \\\hline
         Patch size & $512 \times 512$ \\\hline
         Magnitudes & $> 4$ mb \\\hline
    \end{tabular}
\end{table}

We collected the epicenters of all earthquakes from 2018 until 2021 that were reviewed by ISC. We kept the ones defined as \textit{known} events. We retained only the earthquakes that could create some visible effects, filtering the ones with mb magnitude greater than 4.
We queried the Copernicus system of the European Space Agency to collect samples from Sentinel-1. 
For each epicenter longitude-latitude coordinates $(E_{lg}, E_{lt})$, we collected the area from $(E_{lg} - 0.1, E_{lt} + 0.1)$ to $(E_{lg} + 0.1, E_{lt} - 0.1)$. This creates an area around the epicenter of $\approx 20km \times 20km$. We have also constrained the image selection to IW mode samples.
We collected two samples for each earthquake, one before (we call it ``pre-event'') and one after (we call it ``post-event'') the event. Since the revisit time of Sentinel-1 is 12 days, we created a temporal window of 13 days before and 13 days after the earthquake. We then selected the two images temporally nearest the event, excluding the day of the incident. 
We have excluded epicenters for which we do not have all the two samples.
Furthermore, we collected an extra sample of the same area in a temporal window from 25 days before to 13 days before the earthquake, selecting the furthest day from the ``pre-event'' sample to avoid excessive similarities (we call it ``neutral''). The availability of the third image for each earthquake helps balance the data distribution when considering the tasks based on the bitemporal time series in the following section. 

\begin{figure}[htb]
    \centering
    \includegraphics{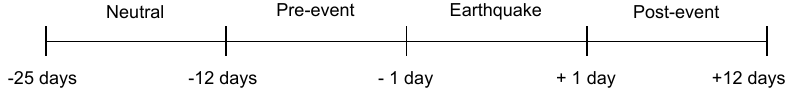}
    \caption{Temporal windows of collected samples}
    \label{fig:timeline}
\end{figure}

\Cref{fig:timeline} depicts the four temporal windows considered for each earthquake. To summarize, three Sentinel-1 images are collected for each earthquake: one in the post-event, one in the pre-event, and one in the neutral temporal window.

Each sample was preprocessed using SAR backscatter with the sigma0 ellipsoid coefficient.
In this way, we collected 155 different events. For each earthquake, we provide three images (neutral, pre-event, post-event), the magnitude expressed in mb, the epicenter coordinates, and the depth (used to identify the hypocenter), when available. 

\begin{figure}[htb]
    \centering
    \includegraphics[width=\linewidth]{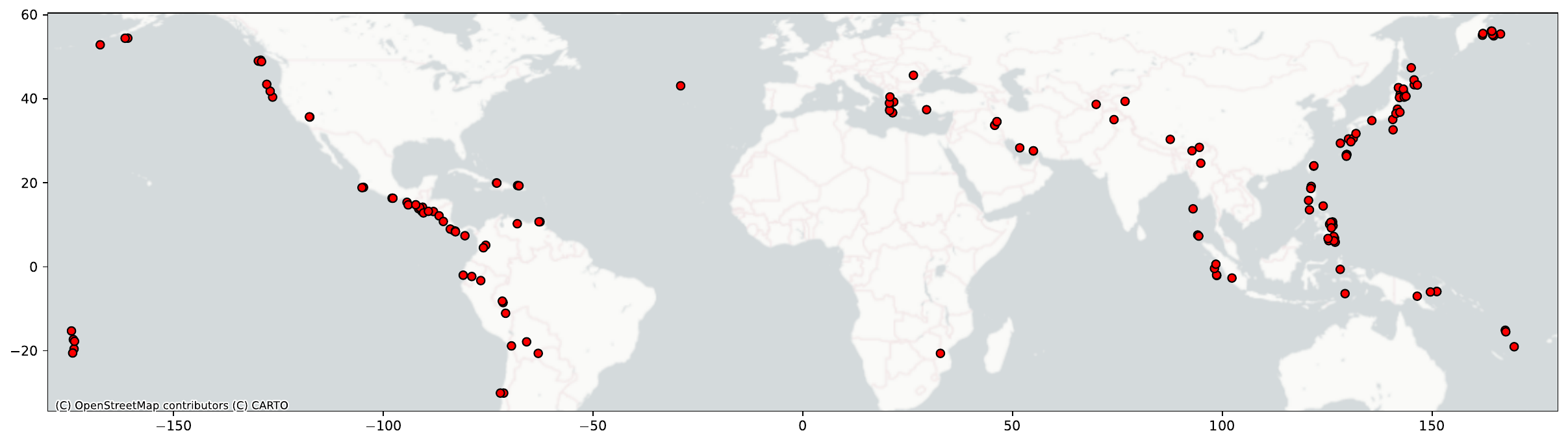}
    \caption{Earthquakes epicenters around the globe}
    \label{fig:earthquakes_map}
\end{figure}

In \Cref{fig:earthquakes_map}, it is possible to see geographical distribution around the globe. Americas, East Asia, and Oceania are best represented, while Africa has the fewest samples. The identified epicenters cover coastal areas, hinterlands, and islands. 

\begin{figure}[htb]
    \centering
    \subfloat[][Magnitudes distribution \label{fig:magnitudes}]{\includegraphics[width=0.5\linewidth]{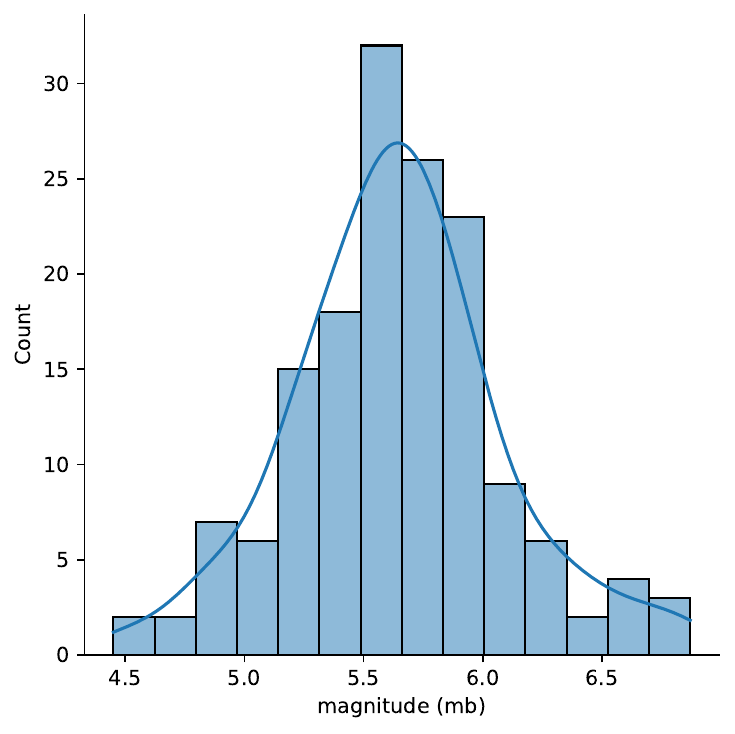}}
    \subfloat[][Magnitude distribution over splits \label{fig:magnitude_split}]{\includegraphics[width=0.5\linewidth]{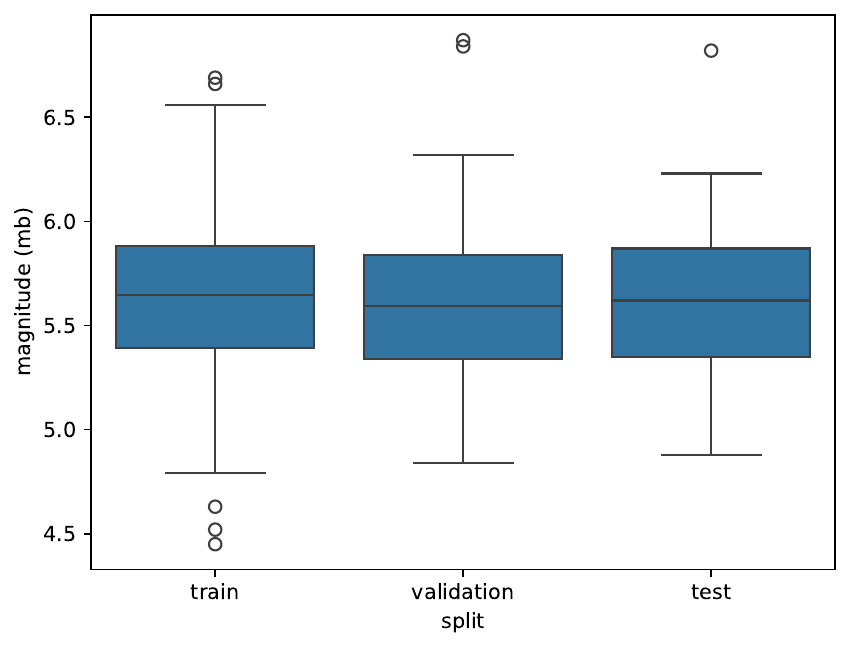}}
    \caption{Magnitude distributions}
\end{figure}

In \Cref{fig:magnitudes}, you can see the distribution of the magnitudes. We collected sufficient samples to approximate the distribution using a normal with $\mu = 5.64, \sigma = 0.44$. The collected values can be considered non-saturated according to \cite{magnitude_saturation}, which poses the upper limit of the scale to $mb_{sat} \approx 8.1$. 

We patched the samples from the center to obtain images of size $512 \times 512 \times 2$, which are more tractable by machine learning models. In this way, we created 1906 patches. We reported a dataset summary in \Cref{tab:dataset_summary}, highlighting the main features.

\subsection{Train, Validation and Test splits}
To create representative splits, we divided the dataset into 70\%-15\%-15\% sets for train, validation, and testing, respectively. 
We decided to create the splits according to the distribution of the magnitudes. All the images (neutral, pre-event, and post-event) associated with the same earthquake event are in the same set/split.
We sample from the original dataset with a uniform distribution. We evaluated the magnitude distribution using the two-sample Kolmogorov-Smirnov test to ensure similar distributions between train-validation and train-test. We obtain a p-value of 0.985 between train-validation and 0.995 between train-test sets. The obtained distributions in \Cref{fig:magnitude_split} highlight the similarities between the three sets in terms of quantiles.

\subsection{Tasks}
Several tasks related to earthquake monitoring can be solved using the data we collected, and we are sharing. Among the others, we suggest using our data to handle the following emergency earthquake-related tasks:

\Crefname{enumi}{Task}{Tasks}

\begin{enumerate}
    \item Earthquake detection with bitemporal image time series \label{task:bi_detection}
    \item Magnitude regression with bitemporal image time series
    \label{task:bi_regression}
    \item Hypocenter or epicenter regression with bitemporal image time series
    \label{task:bi_epicenter}
    \item Change detection with bitemporal image time series
    \label{task:bi_change}
    \item Earthquake detection with a single image \label{task:detection}
    \item Magnitude regression with a single image
    \label{task:regression}
    \item Hypocenter or epicenter regression with a single image
    \label{task:epicenter}
\end{enumerate}

In the following, we formalize each type of task regardless of the input image time series size (the input can be either a single image or a time series composed of two consecutive images).

\subsubsection{Earthquake Detection}
The earthquake detection tasks (\Cref{task:detection,task:bi_detection}) can be formulated as follows. We have a set of time-series $\mathcal{T}$, where each time-series $T_s\in \mathcal{T}$ is composed of $N$ images, with $N \leq 2$, of size $W \times H \times C$ related to the same spatial area $S$ at timesteps $\{T_1,\dots,T_N\}$ and a ground truth value $G_t \in \{0, 1\}$ (where $1$ indicate $S$ was affected by an earthquake, $0$ otherwise). 
Given a training dataset $D_{tr}$, composed of a set of pairs $(T_s, G_t)$, we train a machine learning model $M$. Given a test dataset $D_{ts}$, composed of a set of $T_s$, we can predict $G_t$ for each sample $T_s$ using $M$. This can be framed as a supervised classification task.

\subsubsection{Magnitude Regression}
The magnitude regression tasks (\Cref{task:bi_regression,task:regression}) can be formulated as follows. 
We have a set of time-series $\mathcal{T}$, where each time-series $T_s\in \mathcal{T}$ is composed of $N$ images, with $N \leq 2$, of size $W \times H \times C$ related to the same spatial area $S$ at different timesteps $\{T_1,\dots,T_N\}$ and a ground truth value $G_t \in \{0...M_m\}$ (where $M_m$ is the maximum value for the given magnitude scale). Given a training dataset $D_{tr}$, composed of a set of pairs $(T_s, G_t)$, we train a machine learning model $M$. Given a test dataset $D_{ts}$, composed of a set of $T_s$, we can regress $G_t$ for each sample $T_s$ using $M$.
This regression task is a supervised one.

\subsubsection{Epicenter or Hypocenter regression}
The Epicenter or Hypocenter regression tasks (\Cref{task:epicenter,task:bi_epicenter}) can be formulated as follows. 
We have a set of time-series $\mathcal{T}$, where each time-series $T_s\in \mathcal{T}$ is composed of $N$ images, with $N \leq 2$, of size $W \times H \times C$ related to the same spatial area $S$ at different timesteps $\{T_1,\dots,T_N\}$ and a ground truth composed by the coordinates $(x, y, d)$ (where $x$ and $y$ are the epicenter coordinates and $d$ is the depth of the hypocenter) and a binary label $G_t \in \{0, 1\}$ ($1$ if the area $S$ contains the hypocenter of the earthquake, otherwise $0$). 
Given a training dataset $D_{tr}$, composed of a set of tuples $(T_s, G_t, x, y, d)$, we train a machine learning model $M$. Given a test dataset $D_{ts}$, composed of a set of $T_s$, we can predict $(G_t, x, y, d)$ for each sample $T_s$ using $M$.
If $S$ is affected by an earthquake but does not contain the hypocenter, it should recognize that the tuple $(x, y, d)$ cannot be determined. 
The task is supervised and combines classification (presence or absence of the epicenter) and regression (coordinates of the hypocenter).

\subsubsection{Change Detection}
Change detection (\Cref{task:bi_change}) can be formulated as follows. 
We have a set of time-series $\mathcal{T}$, where each time-series $T_s\in \mathcal{T}$ is composed of 2 images of size $W \times H \times C$ related to the same spatial area $S$ at different timesteps $\{T_1, T_2\}$. We want to generate a binary mask $M_b$ of size $W \times H$ for each $T_s$. The value $1$ indicates the pixel is affected by a substantial change; otherwise, it is set to $0$. The task is an unsupervised semantic segmentation task.
Given a training dataset $D_{tr}$, composed of a set of $T_s$, we train a machine learning model $M$. Given a test dataset $D_{ts}$, composed of a set of $T_s$, we can generate $M_b$ for each sample $T_s$ using $M$.

The previous formulations are general, and the mentioned tasks can be solved using single images (degenerate time series) or bi-temporal time series composed of two images.  

\section{Experiments}
In this section, we present the results obtained for a subset of the proposed tasks, which are \Cref{task:bi_detection,task:regression,task:bi_regression,task:detection}. We have solved the most promising tasks to evaluate the quality of the collected data and the feasibility of the more relevant proposed tasks.

We employed four deep-learning models designed for low-resource devices: MobileNetV2 \parencite{mobilenetv2} and ConvNextV2-Atto \parencite{convnext}, which are convolutional neural networks (CNN), and MiT-B0 \parencite{segformer} and MobileViTV2-1.0 \parencite{mobilevitv2}, which are vision transformers (ViT). We also considered some classical shallow learning models: Support Vector Machines (SVM) with 3rd-grade polynomial and Radial Basis Function kernels and Random Forest (RF). We have chosen these solutions because SVMs can exploit non-linear relations, while RFs, being an ensemble, are more robust.

\subsection{Experimental Settings}
This section reports all settings for the employed models and the input datasets. 

\subsubsection{Input data}
\paragraph{Single image.}
In this setting, the input data is a collection of single images (i.e., a collection of degenerate time series, each one composed of a single image).
We use all the post-event images (as representative of areas already affected by an earthquake) and all the pre-event ones (as representative of areas unaffected by an earthquake). This corresponds to a balanced distribution of the two classes under analysis (affected/unaffected by an earthquake).
In this setting, we do not consider the neutral images (see \Cref{fig:timeline}). Each image has shape $512 \times 512 \times 2$.  

\paragraph{Bi-temporal time series.}
This second setting considers as input a collection of bitemporal time series of images.
When dealing with bitemporal time series of images, we use samples of shape $512 \times 512 \times 4$ concatenating the two components of the time series of shape $512 \times 512 \times 2$ along the channel axis (this approach allows using machine learning models not designed for time series). Each time series is related to a specific earthquake and comprises a pre-event and a post-event image (example of affected area class object) or a neutral and a pre-event sample (example of unaffected area class object). This ensures a balanced distribution of positive and negative examples for the two classes. 

\subsubsection{Models}
\paragraph{Deep Learning Models.} We trained convolutional neural networks for six epochs while vision transformers for ten epochs. This was done due to the slower learning of transformers. The batch size is 16. We used an AdamW optimizer with a cosine-annealing learning rate scheduler with a warmup of 0.1 of the total training steps. The learning rate starts from 0.0001 for CNNs and from 0.001 for ViTs. All models are initialized without any pre-trained weights. The loss is Mean Squared Error (MSE) for regression tasks and Cross-Entropy loss for classification.

\paragraph{Shallow Models.} We trained the Support Vector Machines with 3rd-degree polynomial and RBF kernels. Due to the high number of features, we applied PCA before using classical models. In this way, the size of the features of each sample is reduced to $\approx 2000$.

\subsubsection{Metrics}
We analyzed the results in terms of accuracy for classification tasks since the class distribution is balanced ($\approx 50\%$ of positive and negative samples). Accuracy is defined as:
\begin{equation}
    Accuracy = \frac{CC}{N}
\end{equation}
where $CC$ is the number of samples predicted correctly, and $N$ is the number of samples we asked for a prediction.

We analyzed the results regarding Mean Absolute Error (MAE) for the regression tasks. Still, we also evaluated the accuracy in identifying the occurrence of an earthquake thresholding the predicted magnitude (if magnitude $< 1$, then no earthquake occurred; otherwise, an event affected the area). 
Mean Absolute Error is defined as:
\begin{equation}
    \text{MAE} = \frac{1}{N}\sum_i^N | y_i - \hat{y_i} |
\end{equation}
where $N$ is the number of samples, $y_i$ is the ground truth magnitude, and $\hat{y_i}$ is the predicted magnitude for the $i\text{th}$ sample.

We are also interested in the resource consumption of the presented models, so we report the number of parameters for deep learning models, inference time in seconds, and Mega Floating Points Operation Per seconds (MFLOPs) computed on an Intel(R) Core(TM) i9-10980XE CPU.

\subsection{Earthquake detection with bitemporal time series}
We report in \Cref{tab:detection_results} the results for \Cref{task:bi_detection}. The Random Forest Classifier is the best shallow model in terms of accuracy, providing reasonably high accuracy with a few FLOPs. 

Regarding deep learning models, MobileViTV2 struggles to compete with the others, obtaining $-20\%$ accuracy and $2\times$ FLOPs compared to Mit-B0. The best accuracy results are obtained by the CNNs, which are also less resource-hungry than transformers. MobileNet achieves the best result (94.72\%) with fewer FLOPs than the other models. The differences between this model and the RFC classifier regarding FLOPs and accuracy are evident. Hence, deep learning models are needed to achieve high accuracy values at the cost of a more elevated resource consumption (two orders of magnitude more FLOPs than RFC). 

\begin{table}[htb]
    \centering
    \caption{Performance of models for earthquake detection with bitemporal time-series}
    \label{tab:detection_results}    
    \begin{tabular}{@{}l|c|ccc@{}}
\toprule
Model       & Params & Accuracy $\uparrow$ & Time (s) $\downarrow$ & MFLOPs $\downarrow$  \\ \midrule
SVC (RBF kernel)   & -      & 0.6341   & 0.3640 & 1.0486   \\
SVC (Poly kernel)  & -      & 0.5440   & 0.3138 & 1.0486   \\
RFC         & -      & 0.7241   & 0.3130 & 1.0508   \\
MobileNetV2 (CNN) & 2.2M   & 0.9472   & 0.1109 & 207.5949 \\
MiT-B0      (ViT) & 3.4M   & 0.8865   & 0.0909 & 430.7062 \\
ConvNextV2  (CNN)& 3.7M   & 0.9374   & 0.1003 & 373.7727 \\
MobileViTV2 (ViT) & 4.9M   & 0.6536   & 0.1746 & 964.5956 \\ \bottomrule
\end{tabular}
\end{table}

\subsection{Magnitude Regression with bitemporal time series}
In \Cref{tab:magnitude_results}, we report the results obtained for \Cref{task:bi_regression}. 
Classical solutions like Support Vector Regressors and Random Forest Regressor struggle to compete with deep learning models. 
In this case, the SVR with RBF kernel gets the best results in terms of accuracy among shallow machine learning models, despite RFR being the best in MAE. This means SVR can better distinguish between positive and negative samples, while RFR tends to approximate the magnitude better. Another critical factor is the accuracies of the shallow models, which are all near $0.5$, being generally incapable of making distinctions between the classes.

Deep learning models are also more accurate (in terms of accuracy and MAE) for this task.
Deep learning models provide higher accuracies and MAEs, distinguishing affected and unaffected areas. MiT-B0, which achieves worse MAE than CNNs (ConvNextV2 and MobileNetV2), gets an accuracy comparable to ConvNextV2 and MobileNetV2. Hence, Mit-B0, ConvNext, and MobileNet have the same ability to distinguish between earthquake-affected/unaffected areas. However,  ConvNextV2 and MobileNetV2 estimate the magnitude values better.
The performances of MobileViTV2 are not as satisfactory as the other deep-learning models. In this case, MobileNetV2 obtains the best performance under all metrics among the deep learning models.

\begin{table}[htb]
    \centering
    \caption{Performance of models for magnitude regression with bitemporal time-series}
    \label{tab:magnitude_results}
    \begin{tabular}{@{}l|c|cccc@{}}
\toprule
Model       & Params & MAE $\downarrow$   & Accuracy $\uparrow$ & Time (s) $\downarrow$ & MFLOPs $\downarrow$  \\ \midrule
SVR (RBF kernel)   & -      & 2.2368 & 0.5519   & 0.3640 & 1.0486   \\
SVR (Poly kernel)  & -      & 2.6015 & 0.5440   & 0.3138 & 1.0486   \\
RFR         & -      & 1.9930 & 0.5440   & 0.3130 & 1.0508   \\
MobileNetV2 (CNN) & 2.2M   & 0.5456 & 0.9374   & 0.1109 & 207.5949 \\
MiT-B0 (ViT)     & 3.4M   & 0.8496 & 0.9276   & 0.0909 & 430.7062 \\
ConvNextV2 (CNN) & 3.7M   & 0.6494 & 0.9315   & 0.1003 & 373.7727 \\
MobileViTV2 (ViT) & 4.9M   & 1.7612 & 0.7378   & 0.1746 & 964.5956 \\ \bottomrule
\end{tabular}
\end{table}

\subsection{Earthquake detection with a single image}
\Cref{tab:detection_results_post} reports the results for \Cref{task:detection}. 
It shows the models struggle to discriminate between areas affected by earthquakes and the non-affected ones. The benefits in reducing the computation costs compared to the models trained on bitemporal input data are evident only for shallow models ($\approx 0.5\times$ FLOPs than bitemporal settings), while the deep learning models gain is not worth the decrease in terms of accuracy (compare \Cref{tab:detection_results_post} with \Cref{tab:detection_results}). 

The best shallow model to solve \Cref{task:detection} is SVC with RBF kernel, which achieves $0.60$ of accuracy in a few FLOPs.
Transformers get worse results than this model ($-6\%$ in the worst case). MobileNetV2 provides the best results (with $+10\%$ than SVC with RBF kernel), followed by ConvNextV2. 

MobileNetV2 with single images achieves $-24\%$ in terms of accuracy than bitemporal settings (95\%), and the results are comparable to those of Random Forest (RFC) with bitemporal input data (72\%).

A consistent performance decrease using only one image was expected, as shown in the manual analysis of interferograms of \cite{sentinel1_earthquakes} in which the authors highlighted that earthquake detection is feasible by looking at the differences between two phases.

\begin{table}[htb]
    \centering
    \caption{Performance of models for earthquake detection with a single image}
    \label{tab:detection_results_post}    
    \begin{tabular}{@{}l|l|ccc@{}}
\toprule
Model       & Params & Accuracy $\uparrow$ & Time (s) $\downarrow$ & MFLOPs  $\downarrow$ \\ \midrule
SVC (RBF kernel)   & -      & 0.6047   & 0.1488 & 0.524288 \\
SVC (Poly kernel)  & -      & 0.5440   & 0.1472 & 0.524288 \\
RFC         & -      & 0.5186   & 0.1655 & 0.526554 \\
MobileNetV2 (CNN) & 2.2M   & 0.7104   & 0.0862 & 203.3172 \\
MiT-B0 (ViT)     & 3.4M   & 0.5440   & 0.0710 & 424.8801 \\
ConvNextV2 (CNN) & 3.7M   & 0.6419   & 0.0609 & 370.8897 \\
MobileViTV2 (ViT) & 4.9M   & 0.5988   & 0.1171 & 960.3690 \\ \bottomrule
\end{tabular}
\end{table}

\subsection{Magnitude regression with a single image}
The results in \Cref{tab:magnitude_results_post} for \Cref{task:regression} show, as expected from the previous experiments, that the models are not well capable of identifying affected areas using one single image. We can conclude from the reported accuracies that this is common for both shallow and deep learning models. 

The differences between the used models in terms of MAE and accuracy are less evident than before. Shallow models align with previous performances for \Cref{task:bi_regression}. Conversely, deep learning models decrease their mean performance of $-30\%$ in terms of accuracy and $+1.3$ in MAE compared to the bitemporal setting.

MobileNetV2 gets the best results in terms of MAE but does not perform well in distinguishing between affected and unaffected areas. The decrease in accuracy compared to \Cref{task:detection} is $-16\%$.
ConvNextV2 has the best accuracy, while MiT-B0 is the most balanced solution in both metrics. MobileViTV2 struggles to compete with shallow models in both metrics, too. 

\begin{table}[htb]
    \centering
    \caption{Performance of models for magnitude regression with a single image}
    \label{tab:magnitude_results_post}    
    \begin{tabular}{@{}l|l|cccc@{}}
\toprule
Model       & Params & MAE $\downarrow$   & Accuracy $\uparrow$ & Time $\downarrow$  & MFLOPs $\downarrow$  \\ \midrule
SVR (RBF kernel)   & -      & 2.4088 & 0.5519   & 0.1488 & 0.524288 \\
SVR (Poly kernel)  & -      & 2.5960 & 0.5440   & 0.1472 & 0.524288 \\
RFR         & -      & 2.6204 & 0.5440   & 0.1655 & 0.526554 \\
MobileNetV2 (CNN) & 2.2M   & 2.1595 & 0.5734   & 0.0862 & 203.3172 \\
MiT-B0 (ViT)    & 3.4M   & 2.1821 & 0.5832   & 0.0710 & 424.8801 \\
ConvNextV2 (CNN) & 3.7M   & 2.2353 & 0.5851   & 0.0609 & 370.8897 \\
MobileViTV2 (ViT) & 4.9M   & 2.4645 & 0.5440   & 0.1171 & 960.3690 \\ \bottomrule
\end{tabular}
\end{table}

\section{Conclusion and future directions}
Our experimentation shows the promising results machine learning models achieved in analyzing and identifying areas hit by earthquakes. Deep learning models proved to perform better than shallow models. The drawback is they require more computational resources. Designing less resource-intensive networks is an underexplored direction that should be addressed in remote sensing and hazard management. Our experiments show the great benefits of using bi-temporal time series as input for the analyzed tasks. On the contrary, using a single image does not provide satisfactory results both in terms of performance and in terms of reducing resource consumption. In future works, we plan to investigate the remaining proposed tasks and the possibility of reducing the resource consumption of neural networks for these tasks.

\printbibliography[heading=bibliography]

\end{document}